\documentclass[conference]{IEEEtran}
\usepackage[T1]{fontenc}
\usepackage{xcolor}
\usepackage{parskip}
\usepackage{amsmath,amsfonts,amssymb}
\usepackage{centernot,url}
\usepackage[margin=0.75in]{geometry}
\usepackage{bm}
\usepackage{float}
\usepackage{subfigure}
\usepackage{graphicx}
\usepackage{algorithm2e}
\usepackage{algpseudocode}
\usepackage{multicol, blindtext}
\graphicspath{{/home/admin123/Clustering_MD/IEEE_Conf/plots/} }
\usepackage{multicol}
\usepackage{tikz}
\usetikzlibrary{arrows,decorations.pathmorphing,backgrounds,positioning,fit,petri}
\usetikzlibrary{shapes}
\usepackage{amsthm}

\newcommand{\X}{{\mbox{\boldmath $X$}}}

\newtheoremstyle{break}
  {\topsep}{\topsep}%
  {\itshape}{}%
  {\bfseries}{}%
  {\newline}{}%
\theoremstyle{break}
\newtheorem{definition}{Definition}
\theoremstyle{definition}
\ifCLASSINFOpdf
\else
\fi
\begin{document}
\title{Clustering Mixed Datasets Using Homogeneity Analysis with Applications to Big Data}

\author{\IEEEauthorblockN{Rajiv Sambasivan}
\IEEEauthorblockN{Sourish Das}
\IEEEauthorblockA{Chennai Mathematical Institute\\
H1, SIPCOT IT Park, Siruseri\\
Kelambakkam 603103\\
Email: rsambasivan@cmi.ac.in, sourish@cmi.ac.in}

}


%Telephone: (800) 555--1212, Fax: (888) 555--1212}


\maketitle

\begin{abstract}
\noindent Datasets with a mixture of numerical and categorical attributes are routinely encountered in many application domains. In this work we examine an approach to clustering such datasets using homogeneity analysis. Homogeneity analysis determines a euclidean representation of the data. This can be analyzed by leveraging the large body of tools and techniques for data with a euclidean representation. Experiments conducted as part of this study suggest that this approach can be useful in the analysis and exploration of big datasets with a mixture of numerical and categorical attributes.
\end{abstract}


% \fi
\IEEEpeerreviewmaketitle

\section{Introduction and Motivation}
Datasets with a mixture of categorical and numerical attributes are pervasive in applications from business and socio-economic settings. Clustering these datasets is an important activity in their analysis. Techniques to cluster these datasets have been developed by researchers, see for example \cite{hennig2013find}, \cite{ahmad2007k} and \cite{huang1997clustering}. Techniques to cluster mixed datasets either prescribe a probabilistic generative model \cite{vermunt2002latent} or use a dissimilarity measure \cite{Gower71ageneral} to compute a dissimilarity matrix that is then clustered. Each of these approaches have issues that need to be addressed when they are applied to big datasets - datasets with a large number of instances compared to attributes. For example, latent class clustering uses expectation maximization to estimate model parameters. Scaling expectation maximization and latent class clustering to big datasets is non-trivial (see \cite{bradley1998scaling} and \cite{abarda2017divided}). Similarly, dissimilarity based approaches to clustering have to overcome the computational and storage (memory) hurdles associated with computing and storing a large dissimilarity matrix. This study is limited to dissimilarity based approaches to clustering mixed datasets.\\ 
Clustering methods that are based on a euclidean representation of the data have been well studied by researchers over the years. Novel techniques to cluster big datasets using the euclidean distance have been developed in recent times, for example the mini-batch K-Means algorithm \cite{sculley2010web}. Similarly we have a wide range of tools for related tasks like feature extraction, starting with older conventional methods like principal component analysis to more recent methods like random projections \cite{boutsidis2010random} . If we could determine a euclidean representation of the categorical attributes in a mixed dataset then we will be able to leverage these tools and techniques. Representing categorical attributes in a euclidean space has some difficulties. For example in a dataset with a gender attribute, how do we assign a value to the male and female levels? Should the male be assigned a higher value or a lower value? A natural intuition would be that the data and the application context should determine this, but we still need a theory to frame this as a problem and arrive at an optimal representation of the levels. This is precisely what homogeneity analysis \cite{Michailidis98thegifi} provides. Details of the method are provided in section \ref{sec:overview_ha}.\\
In this study we illustrate methods to cluster big datasets with a mixture of categorical and numerical attributes by first determining an optimal euclidean representation of the dataset. We illustrate this on synthetic and real world data. In the synthetic data, the ground truth is known. Experiments revealed that the clustering solution obtained using the homogeneity analysis based representation of the dataset was very close to the ground truth. Validating clustering solutions when the ground truth is unknown is a difficult task as discussed in \cite{jain1988algorithms}[Chapter 4].  The ground truth is usually unknown in most real world datasets, therefore we take recourse to measures that evaluate quality of clustering using quality measures like compactness of clusters, separation of clusters etc. . Experiments with real world data suggest that the proposed method could be very useful in discovering structure in large datasets (see section \ref{sec:AD_results}). If a partitioning approach is used to cluster the large dataset each partition can be analyzed using an appropriate methodology. Large partitions can be reduced to smaller partitions by reapplying the clustering procedure. Small partitions can be analyzed by sophisticated, computationally expensive techniques if required, for example manifold learning. In summary, the representation of the dataset obtained using homogeneity analysis can be very useful in the analysis and exploration of big datasets with a mixture of categorical and numerical attributes.
\section{Problem Context}\label{sec:pc}
We have a large dataset $\mathbb{D}$ with a mixture of numerical and categorical attributes. This dataset has $n$ rows and $p$ attributes (columns). There are $p_n$ numerical attributes and $p_c$ categorical attributes. The set of categorical attributes, with $p_c$ elements, is represented by $\mathbf{J}$. We need to determine a euclidean representation for the categorical variables in the dataset. $\mathbb{D}_c \subseteq \mathbb{D}$, represents the dataset corresponding to the attributes in $\mathbf{J}$. 
\section{Overview of Homogeneity Analysis}\label{sec:overview_ha}

Homogeneity analysis posits that the observed categorical variables have a euclidean representation in a latent (unobserved) euclidean space. The dimensionality of this space, $r$, is a parameter to this procedure. The representation of a row of $\mathbb{D}_c$ in the latent space $\mathbb{R}^r$ is characterized by the following elements:
\begin{enumerate}
\item The true representation of the  row or instance in the latent Euclidean space, $\mathbb{R}^r$.
\item The optimally scaled representation of the row in terms of the observed attributes. This representation uses an optimal real value for the level of each categorical attribute. This is what we seek to learn.  
\item An edge between an object's true representation and its approximation. This edge represents the loss of information due to the categorical nature of the object's attributes.
\end{enumerate}  

Such a representation induces a bipartite graph. The disjoint vertex sets for this graph are the object's true representation and its approximate attribute representation in the latent Euclidean space. This idea is represented in Figure \ref{fig:bipartite-graph}\\
\begin{figure}[h]
\begin{center}
\begin{tikzpicture}[style=thick, scale = 0.4]
\tikzstyle{row}= [circle,draw=black!50,fill=black!20,thick, inner sep=0pt,minimum size=6mm]
\tikzstyle{attrib1l1}= [rectangle,draw=black!50,fill=black!20,thick, inner sep=0pt,minimum size=4mm]
\tikzstyle{attrib1l2}= [draw, diamond,draw=black!50,fill=black!20,thick, inner sep=0pt,minimum size=4mm]
\tikzstyle{attrib2l1}= [draw, star, star points=6, star point ratio=.6, inner sep=0pt,minimum size=4mm]
\tikzstyle{attrib2l2}= [draw, star, star points=4, star point ratio=.6, inner sep=0pt,minimum size=4mm]
\node[row]	(object 1)						{};
\node[row]	(object 2)	[below=of object 1]	{};
\node[row]	(object 3)	[below=of object 2]	{};
\node[row]	(object 4)	[below=of object 3]	{};
\node[row]	(object 5)	[below=of object 4]	{};
\node[row]	(object 6)	[below=of object 5]	{};

\node[attrib1l1]	(atrrib 1 level 1)	[right=of object 2]	{};
\node[attrib1l2]	(atrrib 1 level 2)	[below=of atrrib 1 level 1]	{};
\node[attrib2l1]	(atrrib 2 level 1)	[below=of atrrib 1 level 2]	{};
\node[attrib2l2]	(atrrib 2 level 2)	[below=of atrrib 2 level 1]	{};
\node[black, above, font=\bfseries]	at (object 1.north) {\scriptsize{Data Rows}};
\node[black, above, font = \bfseries]	(cat quant label) [right=of object 1] {\scriptsize {Category Quantifications}};
\node[black, right]	at (atrrib 1 level 1.east) {\scriptsize{Attribute 1, Level 1}};
\node[black, right]	at (atrrib 1 level 2.east) {\scriptsize {Attribute 1, Level 2}};
\node[black, right]	at (atrrib 2 level 1.east) {\scriptsize{Attribute 2, Level 1}};
\node[black, right]	at (atrrib 2 level 2.east) {\scriptsize{Attribute 2, Level 2}};

\draw [-] (object 1.east) -- (atrrib 1 level 1.west);
\draw [-] (object 1.east) -- (atrrib 2 level 1.west);
\draw [-] (object 2.east) -- (atrrib 1 level 1.west);
\draw [-] (object 2.east) -- (atrrib 2 level 1.west);
\draw [-] (object 3.east) -- (atrrib 1 level 2.west);
\draw [-] (object 3.east) -- (atrrib 2 level 2.west);
\draw [-] (object 4.east) -- (atrrib 1 level 1.west);
\draw [-] (object 4.east) -- (atrrib 2 level 2.west);
\draw [-] (object 4.east) -- (atrrib 1 level 1.west);
\draw [-] (object 4.east) -- (atrrib 2 level 2.west);
\draw [-] (object 5.east) -- (atrrib 1 level 2.west);
\draw [-] (object 5.east) -- (atrrib 2 level 1.west);
\draw [-] (object 6.east) -- (atrrib 1 level 1.west);
\draw [-] (object 6.east) -- (atrrib 2 level 2.west);
\end{tikzpicture}
\end{center}
\caption{Representation of a Dataset with Categorical Attributes as a Bipartite Graph}
\label{fig:bipartite-graph}
\end{figure}
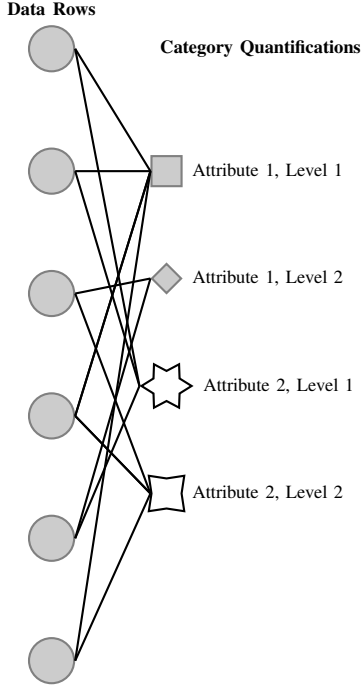
\begin{definition}[Object Score]
The true representation of a row of $\mathbb{D}_c$ in $\mathbb{R}^r$ is called the \textit{object score} of the row. The object scores of $\mathbb{D}_c$ are represented by a matrix $X$ of dimension $n \times r$.
\end{definition}
\begin{definition}[Category Quantification]
A categorical attribute's representation in $\mathbb{R}^r$ is called its \textit{category quantification}. The category quantification for the attributes of $\mathbb{D}_c$ is represented by a matrix $Y$ called the category quantification matrix. The number of category quantifications, $ncc$, is given by:
\begin{equation*}
ncc = \sum_{j\in \mathbf{J}} l_j, 
\end{equation*}
where $l_j$ represents the number of levels for attribute $j$. The dimension of $Y$ is $ncc \times r$.
\end{definition}

The optimally scaled representation of  $\mathbb{D}_c$ in $\mathbb{R}^r$ requires the use of an \textit{indicator matrix}. The indicator matrix, $G$, is a representation of $\mathbb{D}_c$ using a one hot encoding scheme. In a one hot encoding scheme, each attribute of  $\mathbb{D}_c$ is represented by a set of columns corresponding to the number of levels of the attribute. The attribute level taken by the attribute for a particular row is encoded as a $1$, other levels are encoded as $0$. The dimension of $G$ is $n \times ncc$.

Homogeneity Analysis solves the following optimization problem:
\begin{equation}\label{eqn:homals_opt_prob}
\begin{aligned}
& \underset{X,Y}{\text{minimize}}
& & \frac{1}{p_c} \sum_{j = 1}^{j = p_c} \textbf{tr}(X-G_j.Y_j)'(X -G_j.Y_j)\\
& \text{subject to}
& & X^T .X = n.I_r\\
&&& u^T X = 0
\end{aligned}
\end{equation}
Where:
\begin{description}
\item[$\bullet$] $I_r$ is the identity matrix of size $r$
\item[$\bullet$] $u$ is a vector of ones (of length $n$)
\end{description}
The above constraints standardize $X$  and force the solution to be centered around the origin. The constraints also eliminate the trivial solution: $X = 0$ and $Y =0$ . This optimization problem is solved using an Alternating Least Squares (ALS) algorithm. A brief sketch of the steps of the algorithm is provided in Algorithm \ref{proc.ALS-Homals}.\\
\begin{algorithm}
\SetAlgoLined
\SetKwFunction{INITX}{InitializeX}
\SetKwFunction{CANDOX}{CenterAndOrthonormalizeX}
\KwData{$G$}
\KwResult{$X$ and $Y$}
\SetKwData{X}{X}
\SetKwData{Yj}{$Y_j$}
\SetKwData{Dj}{$D_j$}

\X$\leftarrow$ \INITX{}\;
\Dj $\leftarrow G_j'.G_j$\;
\nl\While{solution has not converged}{
\nl\tcc{Minimize $Y_j$ based on current values of $X$. Essentially we are solving $X = G_j.Y_j + \epsilon$ for all $j \in \mathbf{J}$. The solution for this is given below}
\Yj $\leftarrow D_{j}^{-1}G_j'X$\;
\nl\tcc{Now fix $Y_j$ and minimize $X$. The optimal value of $X$ is given below }
\X $\leftarrow p_c^{-1}\sum_{j=1}^{j = p_c}G_jY_j$\;
\nl\tcc{Center and Orthonormalize  X so that the constraints are satisfied. Orthornormalization is performed by an algorithm like Gram Schmidt}
\X $\leftarrow$ \CANDOX{}\; 
}
\caption{Summary of the ALS Algorithm for Homogeneity Analysis}\label{proc.ALS-Homals}
\end{algorithm}
The \textit{loss} associated with the Homogeneity Analysis based solution is the difference between the true representation and the optimally scaled representation. The loss function can be expressed in terms of the attributes as follows:
\begin{align}
\sigma &= \frac{1}{p_c} \sum_{j = 1} ^{j = p_c} SSQ(X - G_j.Y_j)\\ \label{homals:obj-fun2}
	  &= \frac{1}{p_c} \sum_{j = 1}^{j = p_c} \textbf{tr}(X-G_j.Y_j)'(X -G_j.Y_j)
\end{align}
Here,  $SSQ(X - G_j.Y_j)$ refers to sum of the squares of elements of matrix $X - G_j.Y_j$

The Homogeneity Analysis problem can be expressed as an eigenvalue problem (see \cite{Michailidis98thegifi} for details). The loss function of the Homogeneity Analysis solution (Equation \ref{homals:obj-fun2}) can be expressed in terms of the eigenvalues of the average projection matrix $P^*$, for the subspace spanned by the columns of the indicator matrix  $G$ (see \cite{Michailidis98thegifi}). For attribute $j$, the projection matrix is $P_j = G_j .D_j^{-1}.G_j$. The average projection matrix for $p_c$ attributes is given by $P^* = p_c^{-1} \sum_{j = 1}^{j = p_c} P_j$. The loss $\sigma$, for the Homogeneity Analysis solution can be expressed in terms of the eigenvalues of $P^*$. (see \cite{Michailidis98thegifi}):
\begin{equation}\label{eqn:homals-evs}
\sigma = n \left(r - \sum_{s=1}^{s=r} \lambda_s\right),
\end{equation}  
where $\lambda_s$ represents the eigenvalues of $P^*$.
An inspection of Equation \ref{eqn:homals-evs} shows that the number of eigenvalues to use with Homogeneity Analysis solution ($r$ in Equation \ref{eqn:homals-evs}) is a parameter to be chosen. Increasing the number of eigenvalues decreases the loss, however this also increases the dimensionality of the category quantification (each categorical attribute level has a $r$ dimensional representation). In this study we found that using the first eigenvalue ($r = 1$) alone to determine the optimal real valued representation yielded good clustering solutions. This is consistent with the fact that the first eigenvalue holds most information about the attribute's real valued representation. If we use higher values of $r$ (i.e, $r > 1$), then we would replace each categorical value in our original dataset by a $r$ tuple of values.

\section{Application of Homogeneity Analysis to Big Data}\label{sec:aoha}
The original homogeneity analysis (\cite{Michailidis98thegifi}) characterized by Equation \ref{eqn:homals_opt_prob} treats the numerical values in a mixed dataset as categorical variables with a large number of levels or categories. The homogeneity analysis solution determined using such a representation is computable for small or even moderate sized datasets. In big datasets where we could have several hundred thousand unique values of a numeric variable, treatment of numerical variables as categorical variables with a large number of categories introduces computational hurdles associated with processing the matrix $G$. A one hot encoded representation of a variable with a large number of categories can create computational issues. We already have a numerical representation of these variables, therefore we do not include them in the homogeneity analysis computation. Eliminating the numerical variables from the set of variables used for the homogeneity analysis computation implies that the numerical variables do not influence the representation of the categorical variables in the latent space. This is very similar to assumptions made by probabilistic approaches to model mixed datasets like latent class clustering that model the numeric variables and categorical variables independently in the latent space. This is the approach taken in this work. If we are compelled to consider the numerical values in the homogeneity analysis model, we can take a sampling based approach where we determine the euclidean representation based on an appropriate sample, determined for example by stratified sampling.\\

GROUPALS \cite{van1989clusteringn} is a clustering solution that is also based on homogeneity analysis. This approach solves the clustering problem and homogeneity analysis problem together. While this approach is tractable for small and moderate size datasets, it can be impractical with big datasets. The GROUPALS algorithm requires us to provide the number of clusters ($k$) in the data. If this is not known, then we have to compute the GROUPALS solution for a range of $k$ values and then determine the optimal $k$ value by applying a suitable cluster validation index on each solution. We propose an approach that decouples the homogeneity analysis solution and clustering problems. By decoupling the clustering and the homogeneity analysis solution, we compute the homogeneity analysis problem only once. Computing the homogeneity analysis for a range of $k$ values on a large dataset is a time consuming and computationally expensive task. Further, we can apply a wider range of clustering algorithms, not only the k-means like approach used by GROUPALS, to determine an appropriate clustering solution. Similarly it is common solve the clustering and feature selection problems together in small or moderate sized datasets, see \cite{huang1997clustering} and \cite{ahmad2007k}. This again implies that we have to solve the clustering and feature selection problem for a range of $k$ values and then determine the best solution by applying a clustering index to each solution. This is computationally expensive and impractical for big datasets. The scope of this study is limited to datasets that are big in the number of instances rather than the number of attributes. For this study we used incremental PCA \cite{ross2008incremental}, a scalable feature extraction technique, to evaluate feature noise and relevance to the clustering solution. The datasets used in this study had relatively small number of features and feature extraction did not help improve the quality of clustering solutions.
 
\section{Methodology}\label{sec:methodology}
After the euclidean representation of the big dataset has been determined, we need to address the task of clustering it. \cite{jain1988algorithms}[Chapter 2] and \cite{tan2005introduction}[Chapter 8] (more recent) provide a good overview of the algorithmic approaches to clustering. \cite{tan2005introduction}[Chapter 8] provides some guidelines for determining the suitability of a clustering method for a particular application. \cite{jain1988algorithms} mentions that the computational complexity of competing algorithms and the availability of software often dictate the choice of the method. This advice is timeless, it is still applicable today. \cite{bejar2013strategies} provides a summary of the strategies and approaches to clustering big datasets. In this work, we picked three clustering methods for which many good open source implementations exist (such as \cite{scikit-learn} or \cite{r_env}). We illustrate clustering solutions using the mini-batch K-means (\cite{sculley2010web}), the Balanced Iterative Reducing and Clustering using Hierarchies (BIRCH)\cite{zhang1996birch} and the Clustering Large Applications (CLARA) \cite{kaufman2008clustering} algorithms.\\
Validating the developed clustering solutions is the next task. The ground truth in clustering applications is rarely known. For this reason we use a synthetic dataset (described in section \ref{sec:datasets}) to compare the clustering solutions with the ground truth. When the ground truth is known, external clustering indexes \cite{jain1988algorithms}[Chapter 4, section 4.2] are used to validate clustering solutions. When the ground truth is unknown, we need to take recourse to indexes (internal clustering indexes) that measure properties generally observed in good clustering solutions such as good separation and compactness of clusters. The choice of the cluster validation index is again subjective depends on application context. As indicated in \cite{hennig2013find}, there is usually some human intuition about what a good clustering is for a clustering application that helps determine this choice. Further, calculation of these indexes for big datasets is not trivial (see \cite{luna2016approach} and \cite{tlili2014big}) and the availability of reliable software implementations may also play a factor. The adjusted rand index\cite{hubert1985comparing} is a very common choice for comparing a clustering solution with a ground truth. This was used in this work to determine the number of clusters for the synthetic dataset. The Calinski-Harabasz \cite{calinski1974dendrite} index was used to determine the number of clusters for the airline delay dataset (see section \ref{sec:datasets}). This index scales well to large datasets and has been reported to be reliable by researchers over time (see \cite{milligan1985examination} and \cite{arbelaitz2013extensive}).   

\section{Experimental Evaluation} \label{sec:experiments}
\subsection{Datasets}\label{sec:datasets}
The following datasets were used for this study:
\begin{enumerate}
\item \textbf{Synthetic Dataset With Mixed Attributes}: The synthetic dataset used in this study was generated using a two step procedure. The dataset used consists of four attributes - two continuous and two categorical variables with three levels each. The first step of the generating procedure was generate three clusters by sampling a standard isotropic Gaussian distribution. This generates the two continuous variables for the dataset. The second step was to generate the categorical variables for the dataset. Each categorical variable was generated by sampling a multinomial distribution that could generate three category levels. The categorical variables were generated one cluster at a time with a different parameter for each cluster. The multinomial parameter was such that one attribute level was dominant for each categorical attribute in a cluster. The dataset had one million instances.  
\item \textbf{Airline Delay}:This dataset was obtained from the US Department of Transportation's website (\cite{RITA_Delay_Data_Download}). The data represents arrival delays for US domestic flights during January of 2016. This dataset had 11 features and over four hundred and thirty thousand instances. The description of the attributes is provided in Table \ref{tab:jan_21016_delay_data}. 
\begin{table}[ht]
\tiny
\centering
\begin{tabular}{|r|l|l|l|}
  \hline
 & Attribute & Type & Description \\ 
  \hline
1 & DAY\_OF\_MONTH & Ordinal & Day of flight record \\ 
  2 & DAY\_OF\_WEEK & Ordinal & Day of week for
             flight record \\ 
  3 & CARRIER & Nominal & Carrier (Airline) for the flight record \\ 
  4 & ORIGIN & Nominal & Origin airport code \\ 
  5 & DEST & Nominal & Destination airport code \\ 
  6 & DEP\_DELAY & Continuous & Departure Delay in minutes \\ 
  7 & TAXI\_OUT & Continuous & Taxi out time in minutes \\ 
  8 & TAXI\_IN & Continuous & Taxi in time in minutes \\ 
  9 & ARR\_DELAY & Continuous & Arrival delay in minutes \\ 
  10 & CRS\_ELAPSED\_TIME & Continuous & Flight duration \\ 
  11 & NDDT & Continuous & Departure time in minutes from midnight January 1 2016 \\ 
   \hline
\end{tabular}
\caption{Delay Data January 2016}
\label{tab:jan_21016_delay_data} 
\end{table}
\end{enumerate}
The arrival delay attribute is the attribute of interest in this dataset. There are obvious outliers with this attribute, for example it contains flights that actually depart over 24 hours from scheduled time of departure. These obvious outliers were removed by using values corresponding to $99^{th}$ percentile of the values for this attribute. The majority of the outliers remain the dataset used for this study. Details of removing them are described in section \ref{sec:AD_results}. The attributes in the dataset are standardized. 

\subsection{Software Tools}
All modeling for this study was done in \texttt{Python} \cite{Rossum} and \texttt{R} \cite{r_env}. The \texttt{homals} package \cite{homals-pkg} was used for homogeneity analysis. For CLARA, the implementation in the \texttt{fpc} package \cite{fpc-pkg} was used. For BIRCH and mini-batch K-Means, the \texttt{scikit-learn}(\cite{scikit-learn}) implementation was used.

\section{Discussion of Results}\label{sec:dor}

\subsection{Synthetic Dataset}\label{sec:syn_data_results}
The adjusted rand index \cite{hubert1985comparing} was used to compare the clustering results obtained from using the dataset for homogeneity analysis with the ground truth. The adjusted rand index accounts for chance when comparing the ground truth clustering with the clustering solutions produced by the algorithms. The results are shown in Table \ref{tab:ari_syn_data}. The columns in Table \ref{tab:ari_syn_data} represent the adjusted rand index (ARI) obtained with the various values for the number of clusters in the data ($K$).
\begin{table}[H]
\scriptsize
\centering
\begin{tabular}{cccc}
  \hline
 	K & ARI-KM & ARI-BIRCH & ARI-CLARA \\ 
  \hline
	2 & 0.50085 & 0.5565 & 0.7094 \\ 
  	3 &  \textbf{0.94100} &  \textbf{0.9832} & \textbf{0.8713} \\ 
  	4 & 0.7799 & 0.9208 & 0.8502 \\ 
  	5 & 0.7799 & 0.8116 & 0.8394 \\ 
   \hline
\end{tabular}
\caption{Synthetic Dataset - Comparison of Clustering Solutions} \label{tab:ari_syn_data}
\end{table}
\normalsize
For all the algorithms used in this study, the clustering solutions are similar to the ground truth and the adjusted rand index identifies the correct number of clusters in the data.

\subsection{Airline Delay Dataset}\label{sec:AD_results}
As discussed in section \ref{sec:datasets}, the arrival delay attribute is the attribute of interest in the airline delay dataset. This dataset has outliers. This is evident from a review of the quantiles associated with the arrival delay attribute (see Table \ref{tab:Quant_AD}). The $75$ \% of the arrival delay is 5 minutes but the $100^{th}$\% of the arrival delay is 155 minutes. The ground truth for the airline delay dataset is unknown, therefore we use an internal measure of cluster validity, the Calinski -Harabasz index \cite{calinski1974dendrite} to determine the optimal number of clusters. The results of applying clustering algorithms to this dataset after using homogeneity analysis to determine an optimal representation are provided in Table \ref{tab:AD_CHI}.
\begin{table}[H]
\centering
\scriptsize
\begin{tabular}{cccc}
  \hline
 	K & CHI-KM & CHI-BIRCH & CHI-CLARA \\ 
  \hline
	2 & \textbf{100916.2} & 9334.4 & 43527.6 \\ 
  	3 & 94586.8 & \textbf{53220.0} & 83141.3 \\ 
  	4 &  87403.9 & 39539.1 &\textbf{ 90097.6} \\ 
  	5 & 83567.9 & 35304.9 & 76208.3 \\ 
 	6 & 79205.5 & 79205.5 & 78459.6 \\ 
   \hline
\end{tabular}
\caption{Airline Delay Dataset - CHI for Clustering Solutions}\label{tab:AD_CHI}
\end{table}
\normalsize

\begin{table}[ht]
\centering
\begin{tabular}{ccccc}
  \hline
 0\% & 25\% & 50\% & 75\% & 100\% \\ 
  \hline
  -79.00 & -15.00 & -7.00 & 5.00 & 155.00 \\ 
   \hline
\end{tabular}
\caption{Quantiles of Arrival Delay (minutes) - with outliers}\label{tab:Quant_AD}
\end{table}

A review of Table \ref{tab:AD_CHI} shows that the optimal number of clusters for each clustering method is different. Therefore we examine each clustering solution in detail to determine characteristics of each solution. In particular we evaluate the mean arrival delay associated with the clustering solutions. The results are shown in Table \ref{tab:AD_CLUSTER_MEAN_DELAY}. Note that CLARA provides a cluster where the mean delay is nearly three standard deviations from the mean. This is a very useful finding. Indeed, clustering can be used to remove noise from datasets, see \cite{xiong2006enhancing}. Since CLARA provides the most useful clustering at the first level of analysis, we explore this solution further. The profile of the CLARA solution is shown in Table \ref{tab:CLARA_CLUSTER_SUMMARY}. This shows that the clustering produces two large clusters that are characterized by early arrivals. Long flight delays are a rare occurrence and majority of the flights are slightly early, see Table \ref{tab:Quant_AD}. This is consistent with the results we observed with this data. Since we are are applying a partitioning method to cluster the dataset, the clusters produced by CLARA represent a partition of the data. This implies we can analyze each of the clusters in Table  \ref{tab:CLARA_CLUSTER_SUMMARY} independently and arrive at a collective picture of the dataset. Accordingly we applied CLARA clustering to each of the clusters in Table  \ref{tab:CLARA_CLUSTER_SUMMARY}. The results are provided in Table \ref{tab:CLARA_CLUSTER_1_SUMMARY} through Table \ref{tab:CLARA_CLUSTER_3_SUMMARY}. An analysis of these results reveals that the data fall into sub-clusters that either represent early arrivals or slight delays. Table \ref{tab:CLARA_CLUSTER_3_SUMMARY} represents the outliers. Cluster membership associated with a data element can inform us about the likelihood that the data element is associated with a flight delay. For example, membership in sub-cluster 2 of cluster 4 is likely to be associated with delays. In summary, we have been able extract insights by clustering the dataset obtained from homogeneity analysis.
\begin{table}[H]
\centering
\scriptsize
\begin{tabular}{cccc}
  \hline
 Cluster & KMeans & BIRCH & CLARA \\ 
  \hline
    1 & 0.0007 & 0.00030 & -0.2107 \\ 
    2 & 0.0478 & 1.9237 & -0.1937 \\ 
    3 & NA & 0.0478 & \textbf{2.9190} \\ 
    4 & NA & NA & -0.0490 \\ 
   \hline
\end{tabular}
\caption{Airline Delay Dataset - Cluster Mean Delays}\label{tab:AD_CLUSTER_MEAN_DELAY}
\end{table}
\normalsize

\begin{table}[ht]
\centering
\begin{tabular}{cccc}
  \hline
 Cluster & mean & count & sd \\ 
  \hline
	1 & -0.21 & 174810 & 0.59 \\ 
  	2 & -0.19 & 220446 & 0.60 \\ 
  	3 & \textbf{2.92} & 27362 & \textbf{1.19} \\ 
 	4 & -0.05 & 6260 & 0.52 \\ 
   \hline
\end{tabular}
\caption{CLARA - Cluster  Summary} \label{tab:CLARA_CLUSTER_SUMMARY}
\end{table}
\normalsize

\begin{table}[ht]
\centering
\begin{tabular}{cccc}
  \hline
 Cluster & mean & count & sd \\ 
  \hline
	1 & -0.19 & 21776 & 0.58 \\ 
  	2 & -0.21 & 153034 & 0.59 \\ 
   \hline
\end{tabular}
\caption{CLARA - Cluster 1 Summary}\label{tab:CLARA_CLUSTER_1_SUMMARY}
\end{table}

\begin{table}[ht]
\centering
\begin{tabular}{crrr}
  \hline
 Cluster & mean & count & sd \\ 
  \hline
	1 & -0.29 & 29690 & 0.70 \\ 
  	2 & -0.27 & 66828 & 0.50 \\ 
  	3 & 0.22 & 1330 & 0.79 \\ 
  	4 &\textbf{ -0.72} & 1184 & 0.76 \\ 
  	5 & \textbf{0.36} & 21423 & 0.59 \\ 
  	6 & -0.00 & 15691 & 0.59 \\ 
  	7 & -0.23 & 43401 & 0.58 \\ 
  	8 & -0.31 & 40899 & 0.51 \\ 
   \hline
\end{tabular}
\caption{CLARA - Cluster 2 Summary}\label{tab:CLARA_CLUSTER_2_SUMMARY}
\end{table}

\begin{table}[ht]
\centering
\begin{tabular}{crrr}
  \hline
 	Cluster & mean & count & sd \\ 
  \hline
	1 & \textbf{-0.64} & 615 & 0.67 \\ 
  	2 &\textbf{ 0.47} & 612 & 0.84 \\ 
  	3 & -0.04 & 5033 & 0.35 \\ 
   \hline
\end{tabular}
\caption{CLARA - Cluster 4 Summary}\label{tab:CLARA_CLUSTER_4_SUMMARY}
\end{table}

\begin{table}[ht]
\centering
\begin{tabular}{cccc}
  \hline
 	Cluster & mean & count & sd \\ 
  \hline
	1 & 2.04 & 9446 & 0.55 \\ 
  	2 & \textbf{4.34} & 7800 & 0.86 \\ 
  	3 & 2.95 & 936 & 1.08 \\ 
  	4 & 2.51 & 7188 & 0.62 \\ 
  	5 & 3.02 & 1992 & 0.98 \\ 
   \hline
\end{tabular}
\caption{CLARA - Outlier Cluster Summary}\label{tab:CLARA_CLUSTER_3_SUMMARY}
\end{table}

%The dataset that is obtained after removing outliers can be clustered using a similar approach. Such an analysis can be very useful for supervised learning tasks, for example a regression task to predict arrival delays. Clustering applied to the clean dataset may yield clusters that could be used for feature engineering. For example the cluster membership of a data instance could have predictive value in estimating delays. Clustering the large data set can also help us solve the regression task using a divide and conquer approach wherein we develop regression models for each of the clusters.

\section{Conclusion}\label{sec:conclusion}
The original formulation \cite{Michailidis98thegifi} must be carefully applied, accounting for the size of the data and its characteristics (see section \ref{sec:aoha}), when homogeneity analysis is applied to big datasets. Experiments on synthetic data indicate that clustering solutions developed using the euclidean representation determined using homogeneity analysis are similar to the ground truth. Experiments on real datasets indicate that clustering solutions developed using homogeneity analysis can be very useful in analyzing big datasets. Clustering a big dataset with a partitioning based clustering method permits us to apply a divide and conquer strategy for data analysis. The partitions produced by clustering the big dataset can be analyzed independently. When partitions are small, we can consider sophisticated computationally expensive tools for their analysis. In summary, homogeneity analysis can be a useful tool for the exploration and analysis of big datasets with a mixture of continuous and categorical attributes.




%\bibliographystyle{IEEEtran}
\medskip

\bibliographystyle{IEEEtran}

\bibliography{cmdwha}

\end{document}